\documentclass[conference]{IEEEtran}
\ifCLASSINFOpdf
\else
\fi
\usepackage{url}

\usepackage{color}

\usepackage{subfig}
\usepackage{graphicx}
\usepackage{amsmath}
\usepackage{url}
\usepackage{amssymb}

\usepackage{soul} 

\newcommand{\abs}[1]{\left|#1\right|}
\newcommand{\combrule} [1] { #1\hspace{-1.075em}  \bigcirc }
\DeclareMathOperator*{\argmax}{arg\,max}

\begin{document}
%
\title{2CoBel : An Efficient Belief Function Extension for Two-dimensional Continuous Spaces}

\author{\IEEEauthorblockN{Nicola Pellican\`o, Sylvie Le H\'egarat-Mascle, Emanuel Aldea}
\IEEEauthorblockA{SATIE Laboratory, CNRS UMR 8029\\
Paris-Sud University, Paris-Saclay University, Orsay 91405, France\\
Email: \{nicola.pellicano,sylvie.le-hegarat,emanuel.aldea\}@u-psud.fr}
}


%


\maketitle

\begin{abstract}
This paper introduces an innovative approach for handling 2D compound hypotheses within the Belief Function Theory framework. We propose a polygon-based generic representation which relies on polygon clipping operators. This approach allows us to account in the computational cost for the precision of the representation independently of the cardinality of the discernment frame. For the BBA combination and decision making, we propose efficient algorithms which rely on hashes for fast lookup, and on a topological ordering of the focal elements within a directed acyclic graph encoding their interconnections. Additionally, an implementation of the functionalities proposed in this paper is provided as an open source library. Experimental results on a pedestrian localization problem are reported. The experiments show that the solution is accurate and that it fully benefits from the scalability of the 2D search space granularity provided by our representation.
\end{abstract}


%
\IEEEpeerreviewmaketitle

\section{Introduction}
Belief Function Theory (BFT) \cite{dempster2008generalization} \cite{shafer1976mathematical} is  an increasingly  popular framework for the generalization of probability and possibility theory by modeling imprecision and partial ignorance of information, in addition to its uncertainty. BFT is widely used in fundamental tasks which benefit from multi-modal information fusion, such as object detection and tracking \cite{chavez2016multiple} \cite{denoeux2014optimal} , object construction \cite{REKIK2016129}, outdoor localization \cite{ZAIR2017126}, or autonomous robot mapping and tracking \cite{tanzmeister2014grid} \cite{kurdej2014controlling}. Several public evidential theory library exist \cite{gitdempster} \cite{pydempster} \cite{martdempster}, but they are limited to 1D representations. 

The main limitation, when dealing with such theory, since it copes with compound hypotheses, is the size of the set of hypotheses to handle, which may become intractable when the size of the exclusive hypothesis set increases. Such issue becomes critical especially at higher dimensions, as in the case of 2D space hypotheses. Moreover, different tasks may require different levels of precision for the solution, thus calling for a 2D space discretization which would increase quadratically the representation space size. In such a scenario, straightforward binary-word representation of hypotheses, as the one commonly used in 1D evidential theory, which allow bitwise operations and which are therefore very efficient, are no longer possible when the cardinality becomes greater than a few tens of possible solutions. 

For such reasons, some works rely on different approaches to handle the 2D case: by proposing a smart sub-sampling of the 2D space to maintain tractability \cite{ANDRE2015166}; by proposing a sparse representation of the set of hypotheses, and by keeping in memory only the ones which are carrying non-null information \cite{ZAIR2017126}. However, such proposals suffer from several problems which harm their use in practice. Sub-sampling based approaches suffer from  non-scalability, since the operations defined by the framework are still dependent on the size of the frame. They are also precision-bounded, since they involve a coarse approximation of the space. On the other hand, current proposals for sparse representation still suffer from non-unique definition of compound hypotheses, from high accessory management costs, and from the need of non-unique space approximations. 

Following the idea of providing a sparse representation for 2D BFT, and motivated by the great benefit that an efficient representation would carry to high dimensional problems, we propose a new two-dimensional representation which has full scalability properties with respect to the size of the hypothesis space, while allowing a theoretical infinite precision (bounded by the hardware precision limitations). The main contributions of this paper are:
\begin{itemize}
\item The proposal of a new polygon-based compound hypothesis representation, which makes use of polygon clipping operators as basis functions.
\item The use of a hashable representation for fast lookup, and the proposal of a scale independent decision making algorithm. 
\item The release of a public library for multidimensional evidential theory, working with generic representations, and including the proposed definition.
\item The demonstration of the interest of the proposed representation in pedestrian tracking with real data.
\end{itemize}

\section{Settings and definitions}
Let us denote by $\Omega$ the \textit{discernment frame}, i.e. the set of mutually exclusive hypotheses representing the solutions. The power set $2^{\Omega}$ is the set of the $\Omega$ subsets, i.e. the disjunctions of the singleton hypotheses in $\Omega$, having cardinality $2^{\abs{\Omega}}$. The \textit{mass function} $m$, specifying a basic belief assignment (BBA), is defined as $m: 2^{\Omega} \to \left[0,1\right]$ such that $\sum_{A \in 2^{\Omega}}m(A)=1$. A subset of hypotheses $A \in 2^{\Omega}$ such that $m(A)>0$, is a \textit{focal element} of $m$. A BBA is said to be \textit{consonant} if the focal elements are nested: $\forall \left(A,B\right) \in 2^{\Omega} \times 2^{\Omega} \, m(A)>0, m(B)>0 \Rightarrow A \subseteq B \lor B \subseteq A$.

\section{A generic extension and efficient variants}
\subsection{Focal element representation}
\begin{figure*} [t]
\centering
\subfloat[]{\includegraphics[height=5cm]{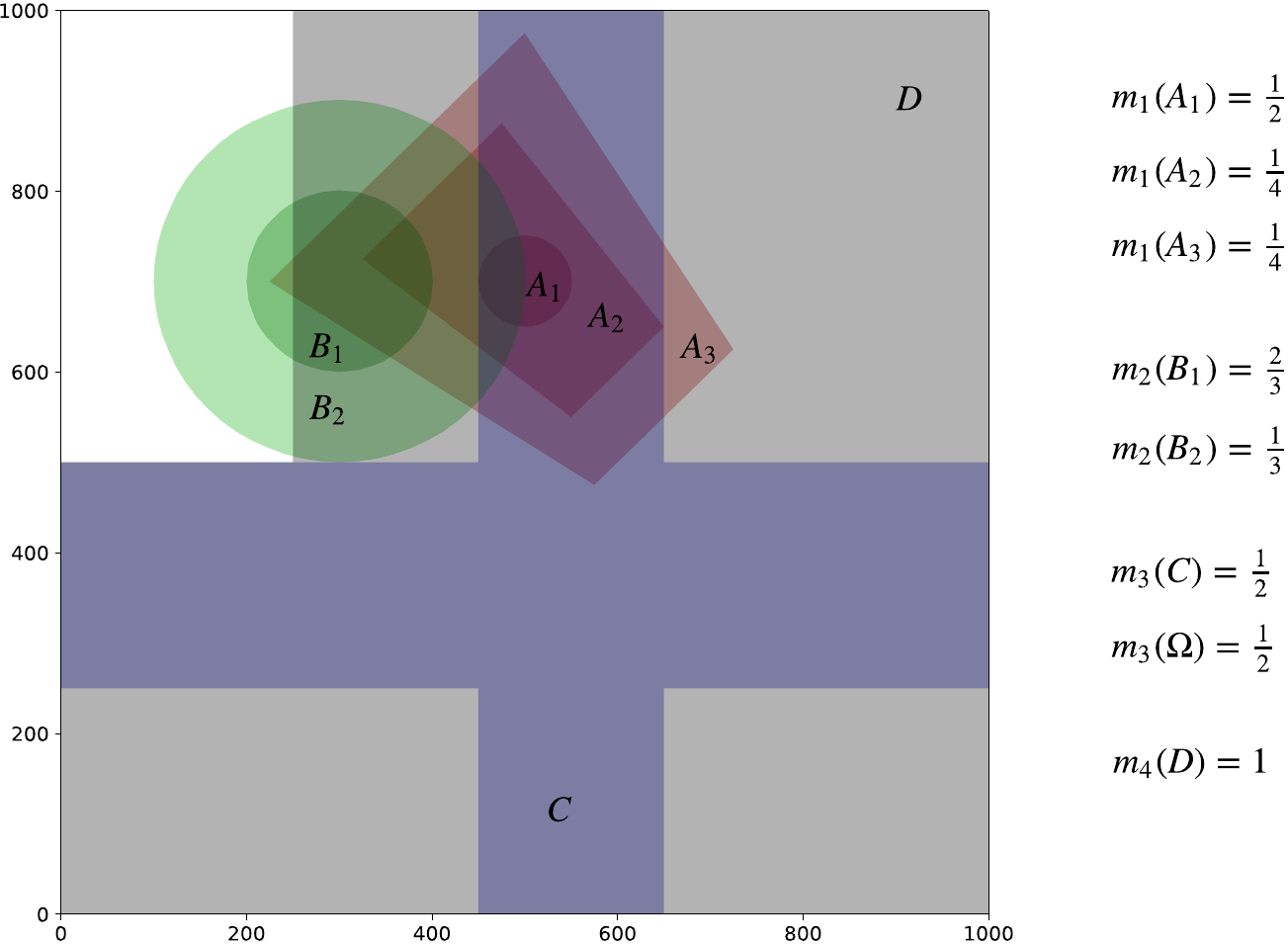} \label{fig:h1}} \; 
\subfloat[]{\includegraphics[height=5cm]{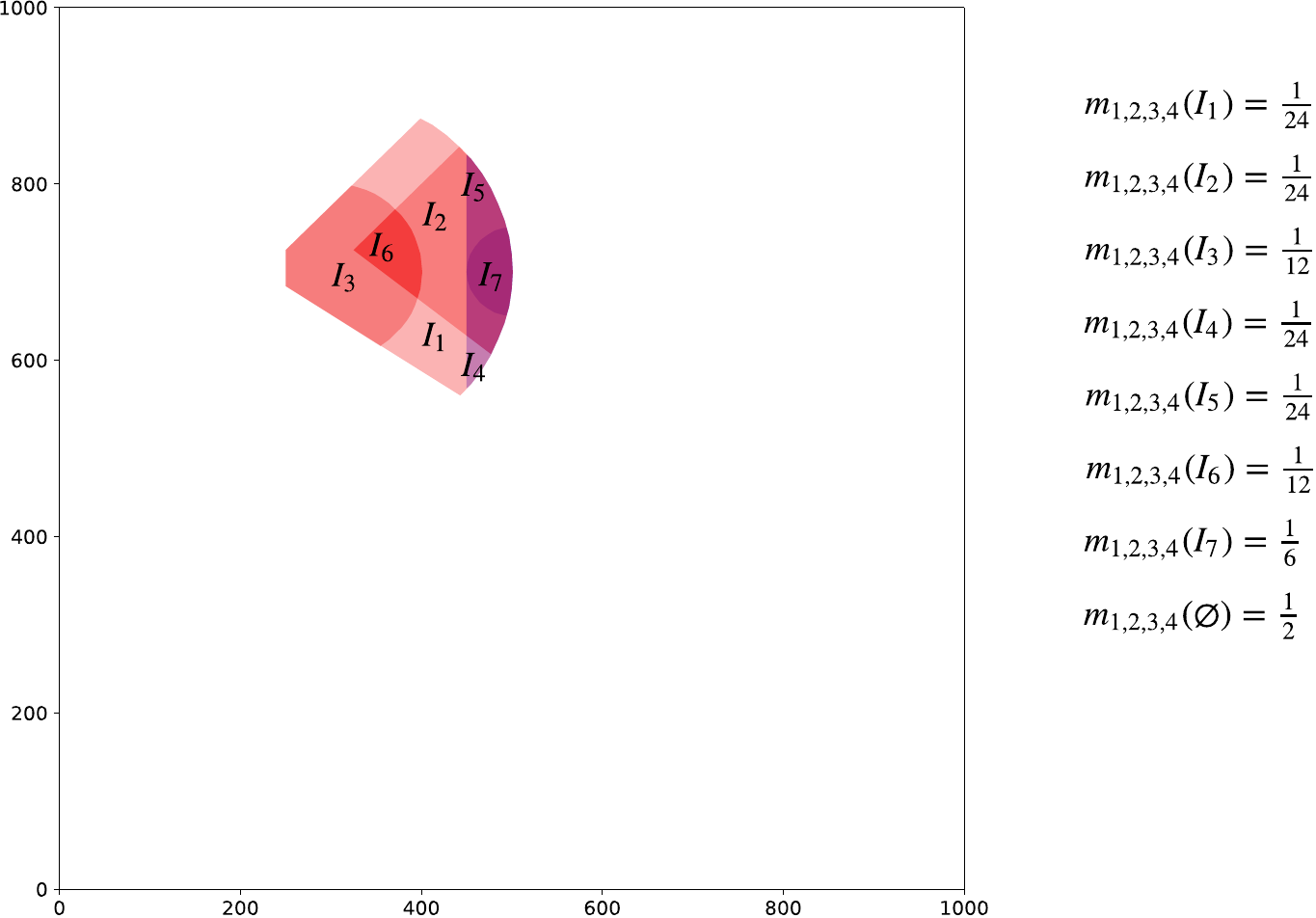} \label{fig:h2}} \\ 
\subfloat[]{\includegraphics[height=4cm]{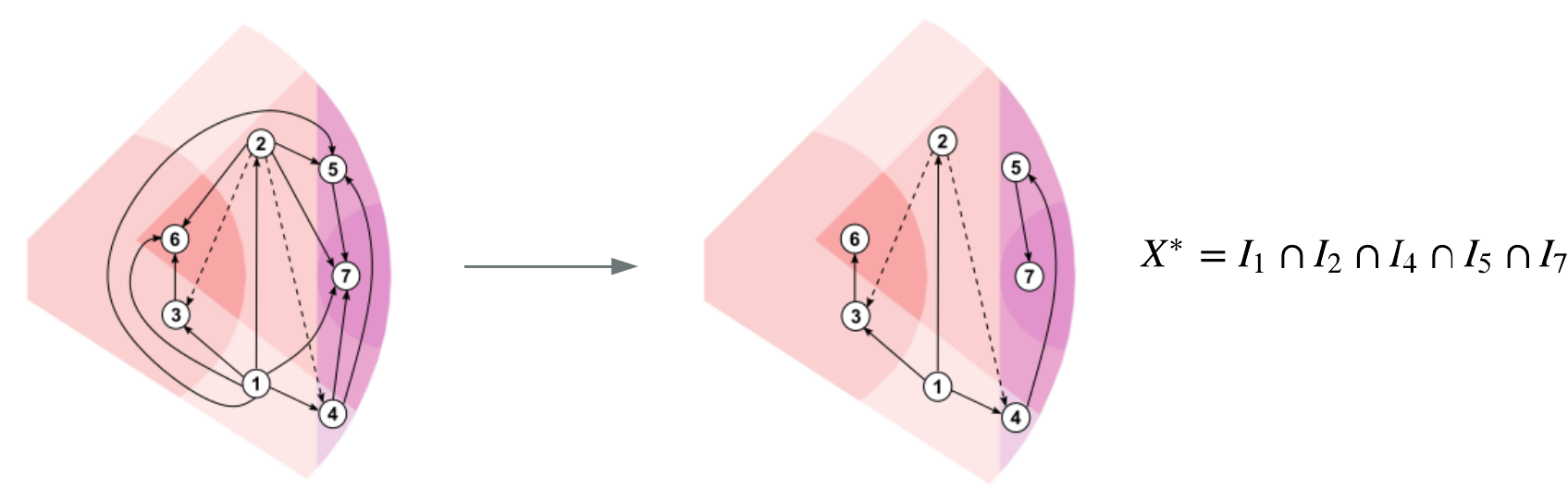} \label{fig:h3}} 
\caption{Toy localization example.  \protect\subref{fig:h1} BBA definition through its focal elements: camera detection (red), track at $t-1$ (green), building presence mask (gray), road presence prior (blue). \protect\subref{fig:h2} Focal elements obtained as a result of performing a conjunctive combination over the defined BBAs. \protect\subref{fig:h3} Intersection graph for $BetP$ maximization and the result of graph simplification. The solid lines show the inclusion relationship, while the dashed lines highlight the intersection relationship. $X^{*}$ is the set with maximum $BetP$ value.}
\label{fig:frames}
\end{figure*}
Let us consider a 2D discernment frame $\Omega$. We will refer, as reference,  to the toy example illustrated in Figure \ref{fig:frames}. Such an example, inspired by \cite{ANDRE2015166}, represents a typical localization scenario, where the discernment frame is a bounded region representing the ground plane.
An exhaustive representation of $\Omega$ discrete hypotheses (as usually) implies a discretization of the area in a grid, where each cell of the grid represents a singleton hypothesis \cite{ANDRE2015166} \cite{ZAIR2017126}. Focal elements are then described by using a binary word, where a bit equal to 1 means that the the cell belongs to the focal set. However, such representation suffers from major drawbacks when used in real world applications. Since there are $2^{\abs{\Omega}}$ potential focal elements, large discernment frames become intractable, when the discretization resolution or the size of the whole area increase. In order to make the representation manageable, \cite{ANDRE2015166} proposes to condition the detections acquired from one sensor in its field of view, and to perform a coarsening at a lower spatial resolution of the focal elements, depending on the physical properties of the sensor. While these workarounds help in practice, they do not make the application fully scalable with the size of the scene, and they involve approximations such as the already cited coarsening, or frequent BBA simplification, which aims at maintaining under control the number of focal elements of the BBAs.

Such limitations derive from the fact that the complexity of any basic operator between focal elements (e.g. intersection, union,...) depends on the cardinality of the focal elements themselves. The works in \cite{REKIK2016129}  overcome this limitation by proposing a representation of any focal element as a set of rectangular boxes, and then by expressing the basic operators as performed on arrays of rectangles. In this setting the complexity of the basic operators will be a function of the number of boxes, but it will be independent from the cardinality of discernment frame. However, such representation suffers from some practical limitations. First, the representation is not unique. The same focal element may be represented by different sets of boxes, which do not allow for fast focal element comparisons and lookup. Second, the box set representation implies a non-unique approximation of the real focal element shape once edges are not parallel to the axes of reference. Geometric approximation of such focal elements may require a very large set of boxes when precision is a concern. Moreover, subsequent operations involve continuous box fragmentation which may be detrimental both for performance and for memory load. In order to avoid deep fragmentation, in \cite{ZAIR2017126} some representation simplification procedures are presented, which in turn increase the cost of BBA management. 

We propose to represent the focal elements as generic polygons (or sets of polygons for focal elements having multiple connected components), by exploiting the capabilities of the generic 2D polygon clipping algorithms in the basic operator implementations. A focal element is represented by a set of closed paths, each of them represented by an ordered array of vertexes (counterclockwise for positive areas, clockwise for holes). We exploit an extension of the Vatti's algorithm for clipping \cite{Vatti} implemented in the Clipper library \cite{clipper}. 

The polygons are constrained to be simple, i.e. defined by closed simple paths (no crossing) and with a minimum number of vertexes (no vertex joining two co-linear edges). Under these constraints, the complexity of the basic operators between two polygons having $n$ and $m$ number of vertexes respectively, is $O(nm)$.
Such lightweight representation presents also the advantages of uniqueness and precision. The (circular) vector of vertexes of a focal element provides a unique representation. The vertex coordinates use integer values for numerical robustness and correctness. This means that the continuous representation provided by polygons implies an underlining discretization. However, differently from the previous approaches, the coordinates can be rescaled at the desired level of precision (up to $\approx 10^{19}$) without any impact on the speed and memory requirements of the algorithm, being bounded only by the numerical representation limits of the hardware. This implies full scalability of the focal elements with respect to their size.
Figure \ref{fig:h1} shows an example of a focal element definition in the case of a localization application. The camera detection (red) is represented as a disk focal element, whereas the focal elements which have the shape of ring sectors embed the imprecision of the location and the ill-knowledge of the camera extrinsic parameters; the track (green) represents the location of the target at the previous frame, whereas its dilation is used in order to model the imprecision in its position introduced by time; the gray and blue focal elements belong to two different BBAs representing scene priors, of building and road presence respectively. The disk shaped focal elements are modeled as 64 to 128 vertexes regular polygons.
\subsection{BBAs combination}
Numerous combination rules exist in order to relate the information provided by two sources. When the sources $m_{1}$ and $m_{2}$ are independent, the conjunctive combination rule is the most popular among them:
$$
\forall A \in 2^{\Omega}, \; m_{1} \combrule \cap \;m_{2} \left( A \right) = \sum_{\substack{(B,C) \mathcal{F}_{1} \times \mathcal{F}_{2}, \\ B \cap C=A}} m_{1}(B)m_{2}(C), 
$$
where $\mathcal{F}_{i}$ is the set of focal elements of $m_{i}$.
In computational terms, the rule involves the construction of a new BBA by performing intersection operations between all pairs of focal elements from the two BBAs. 
According to the sum in the previous equation, when creating a new focal element from an intersection, one has to check for its existence and add up masses if it already exists. Such necessity is not specific of the conjunctive rule, but it is shared with several other rules. 

The above considerations justify the need for a BBA representation which allows for a fast lookup of a focal element in an array. The uniqueness and compactness of the proposed representation allow for an efficient and low collision prone hashing. The sparse set of focal elements of a given BBA can be stored in a hash table, where the circular vector of vertexes is used to compute the hash. For a given polygon, its hash will be unique given a policy to decide the starting vertex (e.g. the top left). The array hashing function is equivalent to the one implemented in the Boost library's \cite{BoostLibrary} \textit{hash\_range} method.

The binary-word representation, in comparison, uses the full word as a unique key. However, the key length (in number of bits) grows linearly with the cardinality of the discernment frame, needing the use of big data structures in order to store it. On the other side, the proposed hash has a fixed length, while having collision resistance property. The box set representation, being not unique, does not allow for direct hashing without the extraction of the minimal set of vertexes on the boundary. A cheap alternative could be to hash the bounding box of the focal element, but this could cause frequent collisions, since it is common to have spatially close focal elements related to the same BBA. 

Figure \ref{fig:h2} illustrates the result of the conjunctive combination of the sources introduced in  Figure \ref{fig:h1}. Seven focal elements are produced.

\subsection{Decision making} \label{betp}
Once different sources have been combined, the decision is generally taken on singleton hypotheses $w$ by maximizing the \textit{pignistic probability}, defined as:
$$
\forall w \in \Omega,\; BetP(w)=\frac{1}{1-m(\emptyset)} \sum_{B \supseteq w} \frac{m(B)}{\abs{B}}.
$$
Even if the search space size is now $\abs{\Omega}$, the decision making process is still dependent on the cardinality of the discernment frame, and thus not scalable, limiting the precision level which can be set for a specific context. 

In order to overcome this limitation, we propose a maximization algorithm which is independent from the cardinality of the sets, and which is only related to the number of focal elements in the BBA. The underlying idea is that, since $BetP$ is an additive measure, its maximum value can be located only in areas of the discernment frame which present \textit{maximal intersections}, defined as follows: given a set of focal elements $\mathcal{A}=\left\lbrace A_{1},\hdots,A_{n}\right\rbrace$, a maximal intersection $I_{m}$ satisfies:
\begin{align*}
&I_{m}=\bigcap_{A_{k} \in \tilde{\mathcal{A}}} A_{k}, \tilde{\mathcal{A}} \subseteq  \mathcal{A}, \abs{I_{m}}>0 \; s.t. \\ &\nexists A_{s} \in \mathcal{A} \setminus \tilde{\mathcal{A}}, \abs{A_{s} \cap I_{m}} >0.
\end{align*}
Finally the set $X^{*}$ of hypotheses that maximizes the $BetP$ is researched within the set of maximal intersections $\mathcal{I}$:
$$
X^{*}=\argmax_{I_{m} \in \mathcal{I}} \frac{BetP\left(I_{m}\right)}{\abs{I_{m}}},
$$
where the $BetP$ function for compound hypotheses derives from the generalized formula:
$$
\forall A \in 2^{\Omega},\; BetP(A)=\frac{1}{1-m(\emptyset)} \sum_{B \in \mathcal{F}, B \cap A \neq \emptyset} \frac{\abs{A \cap B}}{\abs{B}}m(B).
$$
Consequently to this formulation, the $BetP$ maximization algorithm reduces to the subproblem of maximal intersection search. Let us assign an  ordering to the set of focal elements for the given BBA. For optimization reasons explained further, the focal elements are labeled according to decreasing cardinality and the ordering follows the element label. We build a directed acyclic graph (DAG) $G=(V,E)$ where each node $v \in V$ is a focal element and an edge $e \in E$ represents a non empty intersection between two focal elements. The direction of the edges follows the given topological order. Each node is iteratively selected as the root. For each root a depth first search strategy is used to traverse the graph. The graph traversal is performed as follows: given the current node $v_{i}$, the intersection between all the nodes of the current path is propagated as $I_{i}$; given an edge $e=(i,j)$, the node $A_{j}$ is explored if $\abs{I_{j}}=\abs{I_{i} \cap A_{j}}>0$. Such an operation is equivalent to performing a dynamic graph pruning which is a function of the current path. Once a leaf $l$ is reached (a leaf is a node without any edge which can be further explored), the resulting $I_{l}$ is a candidate for maximal intersection. However, it could be non-maximal, as its associated set $\tilde{\mathcal{A}}$ could be a subset of a maximal intersection which has already been found. So, when a maximal intersection $I_{m}$ is found, the list $p_{m}$ of focal sets involving it is stored (using a bit-set representation). Once the new candidate $I_{l}$ is produced, the $p_{l}$ list is tested for inclusion against the stored candidates (by an AND operation between the bit-sets). Even if the number of node visits can be very large in the worst case, in practice, the number of operations is much lower, since the dynamic pruning helps to cut out early dead paths. 

Moreover, further optimization can be performed by inspecting the  inclusion relationships between focal elements. Consider the node $v_{j}$ as the current root. If $A_{j} \subseteq A_{i}$, for some $i<j$, there exists no maximal intersection including $A_{j}$ and not $A_{i}$. This implies that no maximal intersection can be found starting from $v_{j}$ as root.  Thus, only the focal elements not included in others preceding them in topological order are used as root nodes (\textit{root suppression}). This is the reason why we impose the topologically ordered in ascending order of cardinality, since any edge representing inclusion will be directed from the including to the included focal element. This allows us to exploit root suppression as much as possible.

Following the same principle, an \textit{early stopping} criterion can be introduced. Let us consider the algorithm being executed for a root $v_{r}$. Given the current node in a path $v_{i}$ and an edge $e=(v_{i},v_{j})$, the node $v_{j}$ is explored only if it is not a subset of any previous root $v_{k},k<r$. This derives from the fact that since $v_{k}$ is no longer reachable, every path containing $v_{j}$ is non maximal.

Given the mentioned topological order, a \textit{graph simplification}  can be applied to reduce the number of edges in the graph. Given a node $v_{j}$ having more than one incoming connection with a superset focal element, all the inclusion connections but the one from the highest index in topological order can be removed. The reason behind this is that any path which contains $v_{j}$ must contain all its including sets, so, given a list of including nodes $\mathcal{V_{I}}=\left\lbrace v_{I,1},\hdots v_{I,k}\right\rbrace$, a path between $v_{I,n}, n=1\hdots k$ and $v_{j}$ must include all the $v_{I,i},n<i\leq k$, thus only $v_{I,k}$ can have a direct edge to $v_{j}$. Such optimization leads to clear performance gains when inclusion chains are present (such as when dealing with consonant BBAs). An inclusion chain of $k$ elements leads to a complete subgraph in the output DAG, with $2^k$ possible paths. However, after graph simplification, only the edges going from element $A_{i}$ to $A_{i+1}$ are kept, resulting in a single path including all the nodes. 

Figure \ref{fig:h3} shows the intersection graph and its simplification for the proposed toy example. Two intersection graphs are present, and $X^{*}$ is selected as the one at maximum $BetP$. For this example,raw traversal intersection graph performs 42 node visits, while with optimizations 12 are executed. On the other hand, a straightforward $BetP$ maximization by singleton hypothesis exploration would process 1100 locations (included into at least one focal element) with a factor 10 subsampling of the discernment frame.



\section{Experiments}
We present test results on a tracking application scenario, which makes use of the proposed representation on real data, as well as of our publicly available \textit{2CoBel} library, embedding all the described methodologies, and exploited throughout the entire testing.
\subsection{The 2CoBel library}
2CoBel is an open source\footnote{Implementation available at: \\ \url{https://github.com/MOHICANS-project/2CoBel}}  evidential framework embedding essential functionalities for generic BBAs definition, combination and decision making. An \textit{Evidence} object defines common operations for a BBA containing any generic type of \textit{FocalElement}. The current supported methods are: mass to Belief Functions conversion (plausibility, belief, commonality), conjunctive and disjunctive rules, vacuous extension and marginalization, conditioning, discounting, (generalized) $BetP$ computation, $BetP$ maximization (with singleton hypothesis enumeration or maximal intersections). Different types of $FocalElement$ are supported, each of them defining basic operators (intersection, union, equality, inclusion) : \textit{unidimensional} (hashable), representing the 1D focal element as a binary string; \textit{2D bitmap}, providing a bitmap representation as in \cite{ANDRE2015166}; \textit{2D box set}, implementing the definition and focal elements simplification operations proposed in \cite{ZAIR2017126}; \textit{2D polygon} (hashable), implementing our proposed representation. 

The library has full support for cartesian product of discernment frames.


\begin{figure*} [t]
\centering
\subfloat[]{\includegraphics[height=1.5cm]{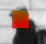} \label{fig:r1}} \; 
\subfloat[]{\includegraphics[height=3.5cm]{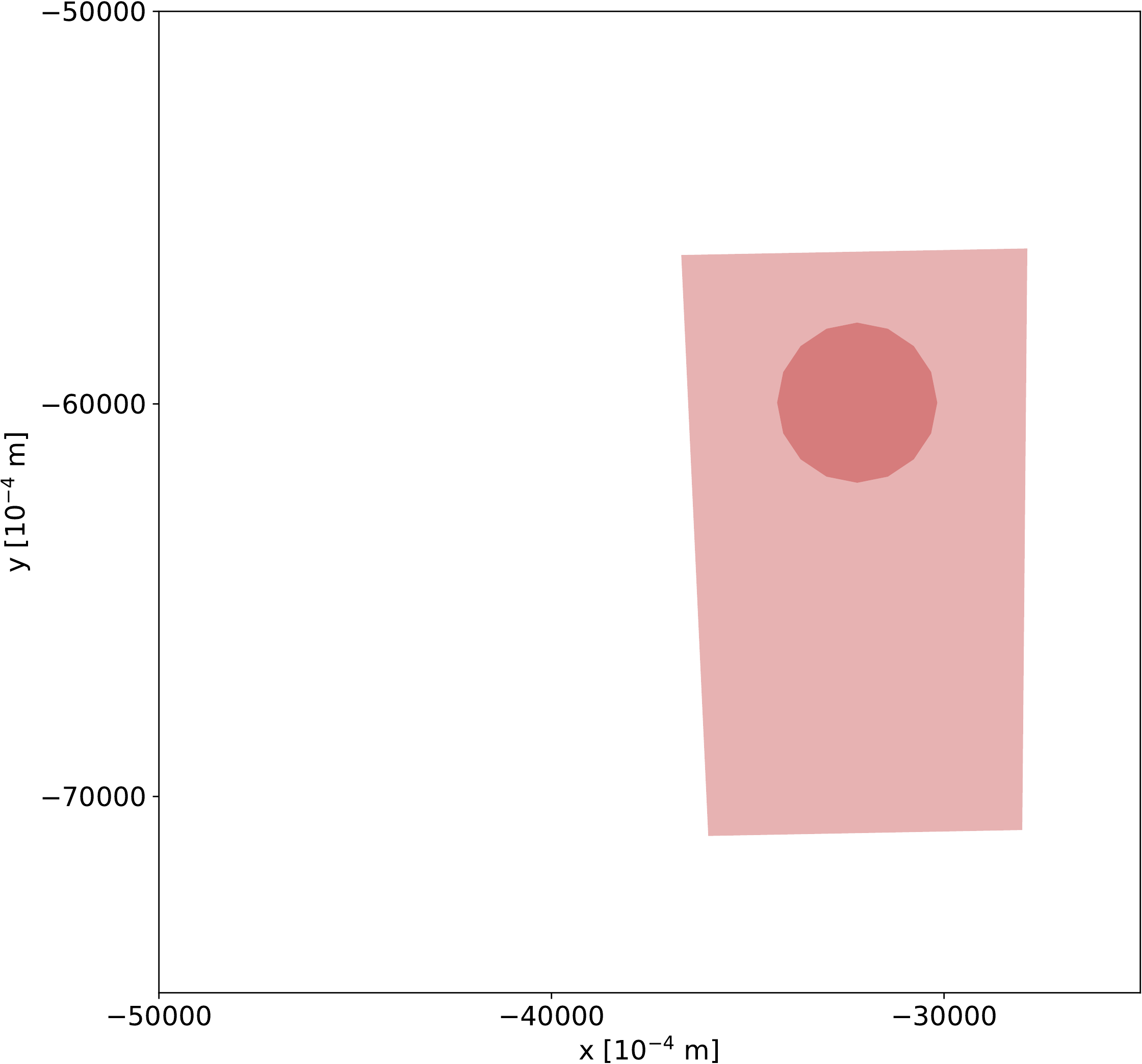} \label{fig:r2}} \;
\subfloat[]{\includegraphics[height=3.5cm]{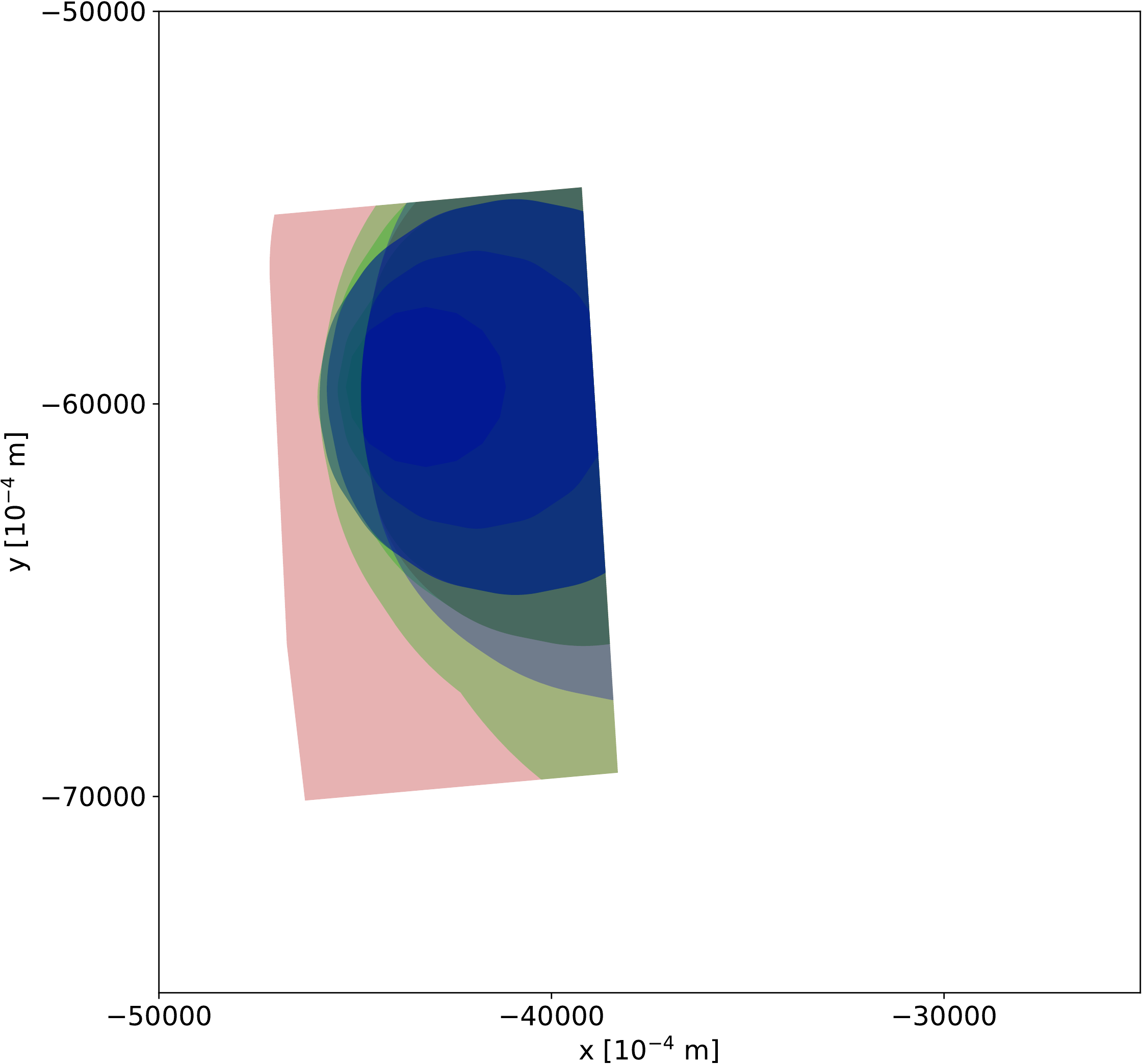} \label{fig:r3}} \;
\subfloat[]{\includegraphics[height=3.5cm]{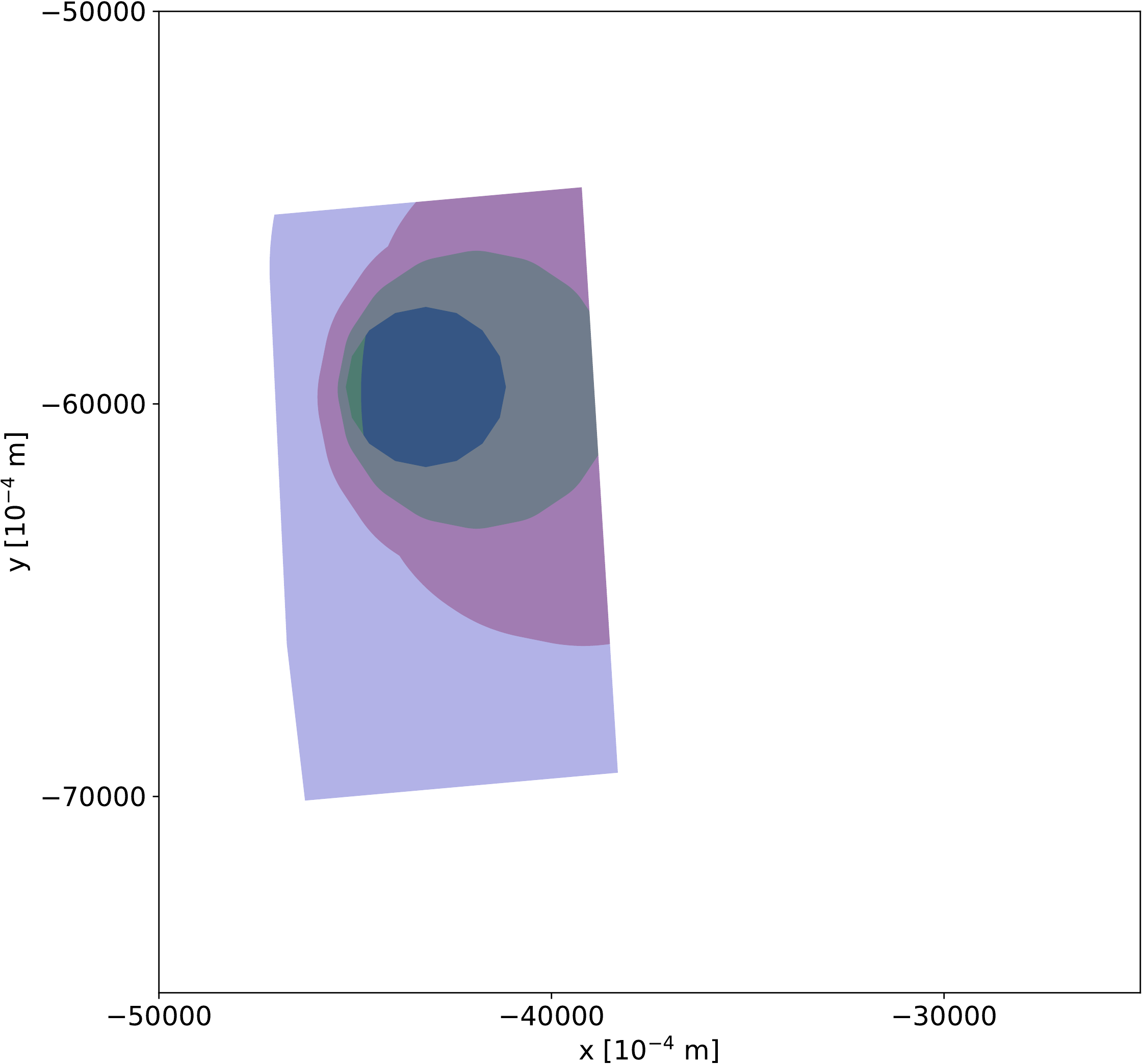} \label{fig:r4}} \;
\subfloat[]{\includegraphics[height=3.5cm]{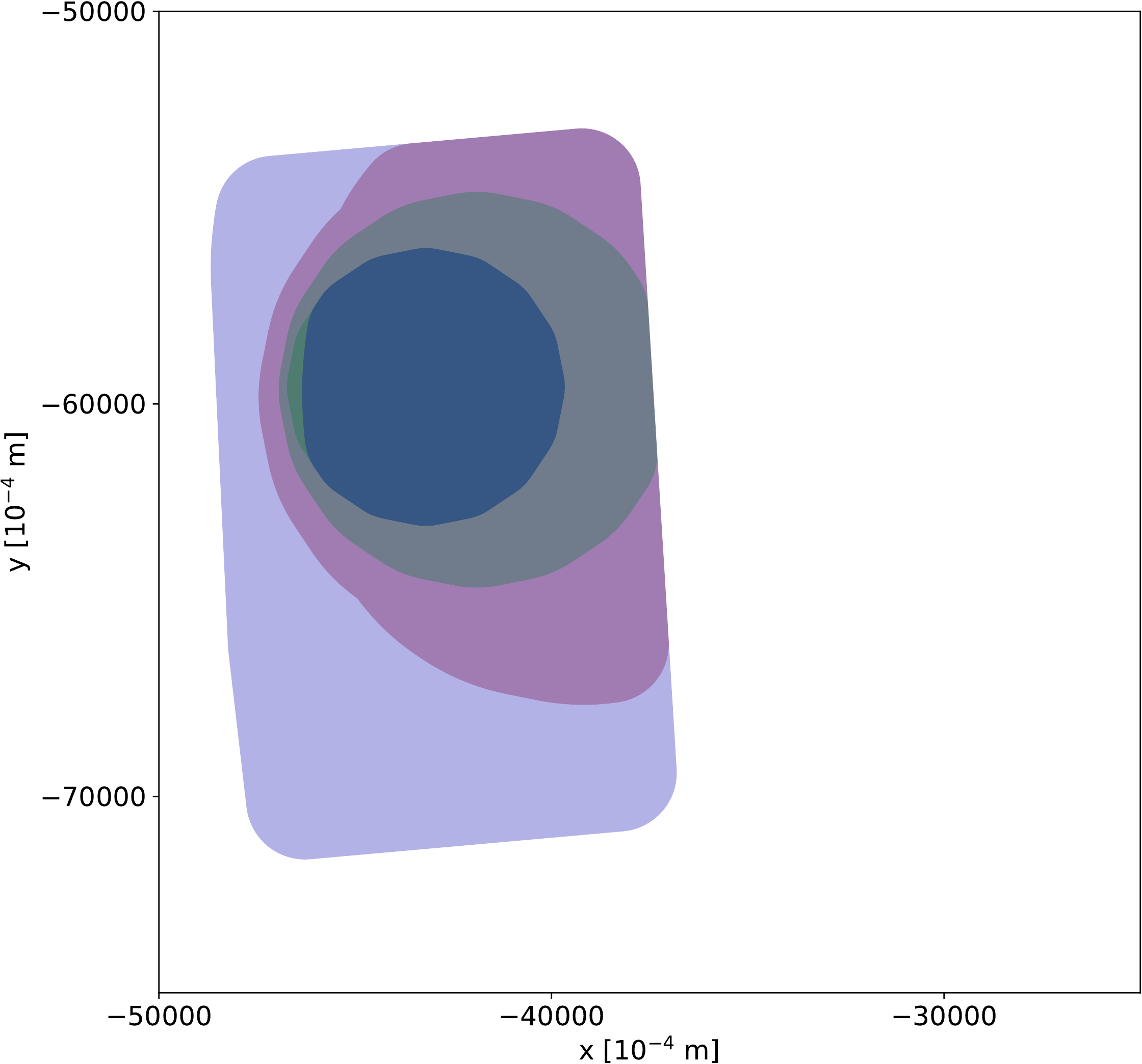} \label{fig:r5}}
\caption{Example of pedestrian tracking steps.  \protect\subref{fig:r1} Pedestrian detection blob. \protect\subref{fig:r2} Focal elements of detection BBA $m_{d_{0}}$ on the ground plane at $t=0$. \protect\subref{fig:r3} Focal elements of the conjunctive combination $\widetilde{m}_{t_{0},7}$  between the track and the associated detection at $t=7$ (16 focal elements). \protect\subref{fig:r4}Focal elements of the BBA simplification of $\widetilde{m}_{t_{0},7}$ with the Jousselme's distance criterion (5 focal elements).  \protect\subref{fig:r4} Focal elements after dilation of the track BBA $m_{t_{0},8}$ by polygon offsetting.}
\label{fig:frames}
\end{figure*}

\subsection{Case study: pedestrian tracking}
We apply the proposed representation to the problem of tracking pedestrians detected by imprecise sensors, on the ground plane. The  belief function framework allows for direct modeling of the imprecision associated with the detections and the tracks and provides a measure for data association between detections and tracks.

We make use of the detector proposed in \cite{pellicano}, which performs low level information fusion from multiple cameras in order to provide a dense pedestrian detection map, together with pedestrian height estimations, in a range between $1.4 \; m$ and $2 \: m$. The output of the detector allows to project and track detections on the ground plane. We demonstrate the use of the \textit{2D polygon} representation provided in the \textit{2CoBel} library in order to perform joint multiple target tracking in the \textit{Sparse} sequence presented in \cite{pellicano}. We perform tracking on the provided detections for 20 frames of the \textit{Sparse} sequence, and we  measure the localization error of the real tracks (13 pedestrians, 4 standing and 9 moving) with respect to the ground truth.
\subsubsection{Discernment frame definition}
The area under analysis is the ground plane region where the field of views of the cameras overlap. The area of the analysis region is $330\;m^2$. The algorithm is run at a resolution of $10^{-4}\;m$, so the cardinality of the discernment frame is $\abs{\Omega}=33\times 10^9$. While the desired localization precision is $10^{-2}\;m$, the chosen resolution is higher for increased robustness to rounding errors. 
\subsubsection{BBA construction and assignment}
Given a detection $d_{i}$ at time $t$ located in $(x_{i},y_{i})$, we build a consonant BBA consisting of two focal elements. The first focal element is a disk centered at $(x_{i},y_{i})$ and with a radius of $20 \; cm$, taking into account the person's head and shoulder occupancy on the ground plane; the second focal element is a ring sector (approximated by a trapezoidal shape), which embeds the height uncertainty (on the direction point towards the camera location) and the camera calibration imprecision. In order to break the symmetry, the two focal elements are not assigned with 0.5 mass each, but with 0.51 for the internal disk and 0.49 for the trapezoid. In the presented case the choice of the mass allocation has a negligible impact on the quantitative results, while it may become critical when additional sensors/sources are included into the problem.

\subsubsection{Data association and combination}
Given a set of tracks at time $\delta$, $\mathcal{T}=\left\lbrace t_{1},\ldots , t_{k} \right\rbrace$ and a set of detections $\mathcal{D}=\left\lbrace d_{1},\ldots , d_{h} \right\rbrace$, the data association aims to compute an optimal one-to-one association set $A_{l}= \left\lbrace (t_{i},d_{j}), i \in \left\lbrace 1\ldots k \right\rbrace , j \in \left\lbrace 1\ldots h \right\rbrace  \right\rbrace$ with respect to some defined cost. A  $(t_{i},\emptyset)$ association means that the track is into an inactive state (so it keeps propagating until it associates with a new detection or dies), while a $(\emptyset,d_{j})$ association means a new track has to be initialized with detection $d_{j}$. We make use of the criterion in \cite{ristic2006tbm} to define the association cost:
$$
C_{t_{i},d_{j}}=- \log \left( 1- m_{t_{i}} \combrule \cap \;m_{d_{j}} ( \emptyset ) \right),
$$
which expresses the data association task as a conflict minimization problem, which can be solved by the use of the Hungarian algorithm.

The data association task is followed by a conjunctive combination which produces for every $(t_{i},d_{j})$ the new track:
$$\widetilde{m}_{t_{i},\delta}=m_{t{i},\delta} \combrule \cap \;m_{d_{j}} \combrule \cap \;m_{p},$$
where $m_{p}$ corresponds to the prior. It performs a masking operation on the visible region of interest of the camera on the ground plane.

\begin{figure*} [t]
\centering
\subfloat[]{\includegraphics[height=6cm]{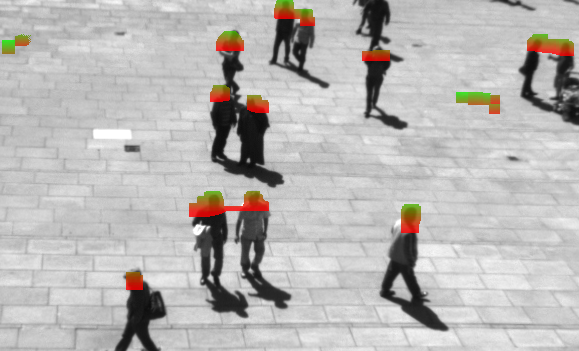} \label{fig:s1}} \\ 
\subfloat[]{\includegraphics[height=6.2cm]{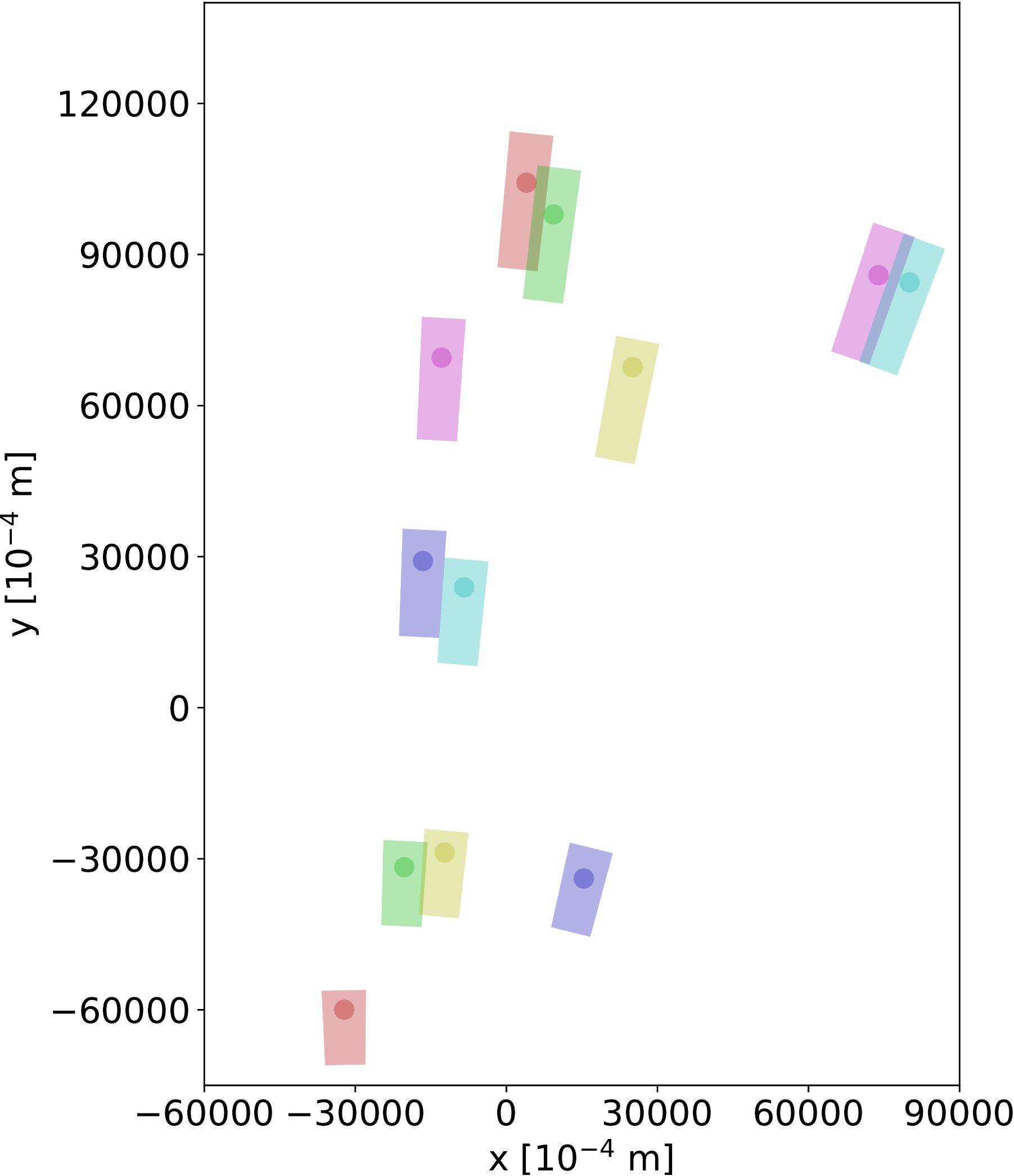} \label{fig:s2}} \; 
\subfloat[]{\includegraphics[height=6.2cm]{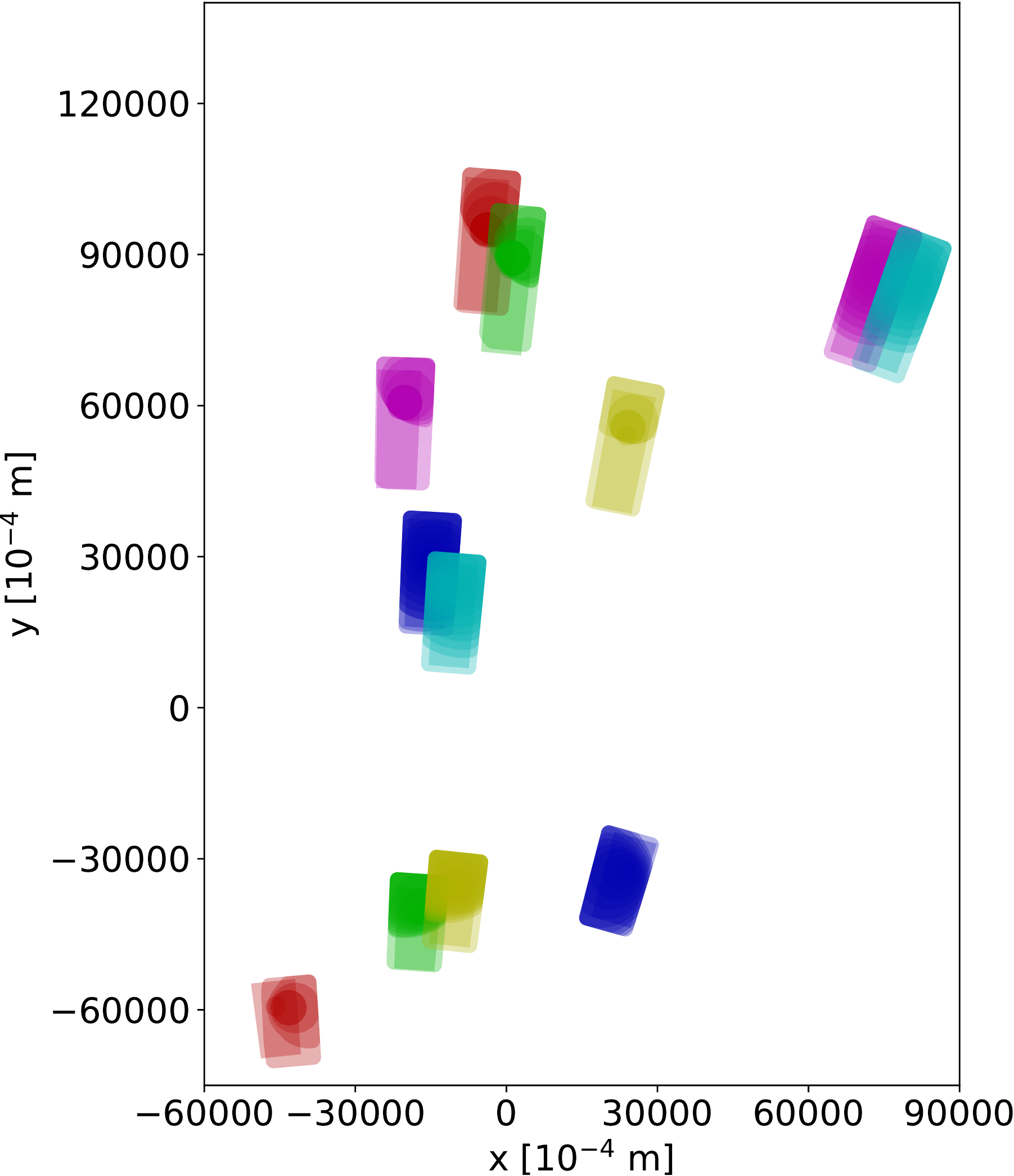} \label{fig:s3}} \;
\subfloat[]{\includegraphics[height=6.2cm]{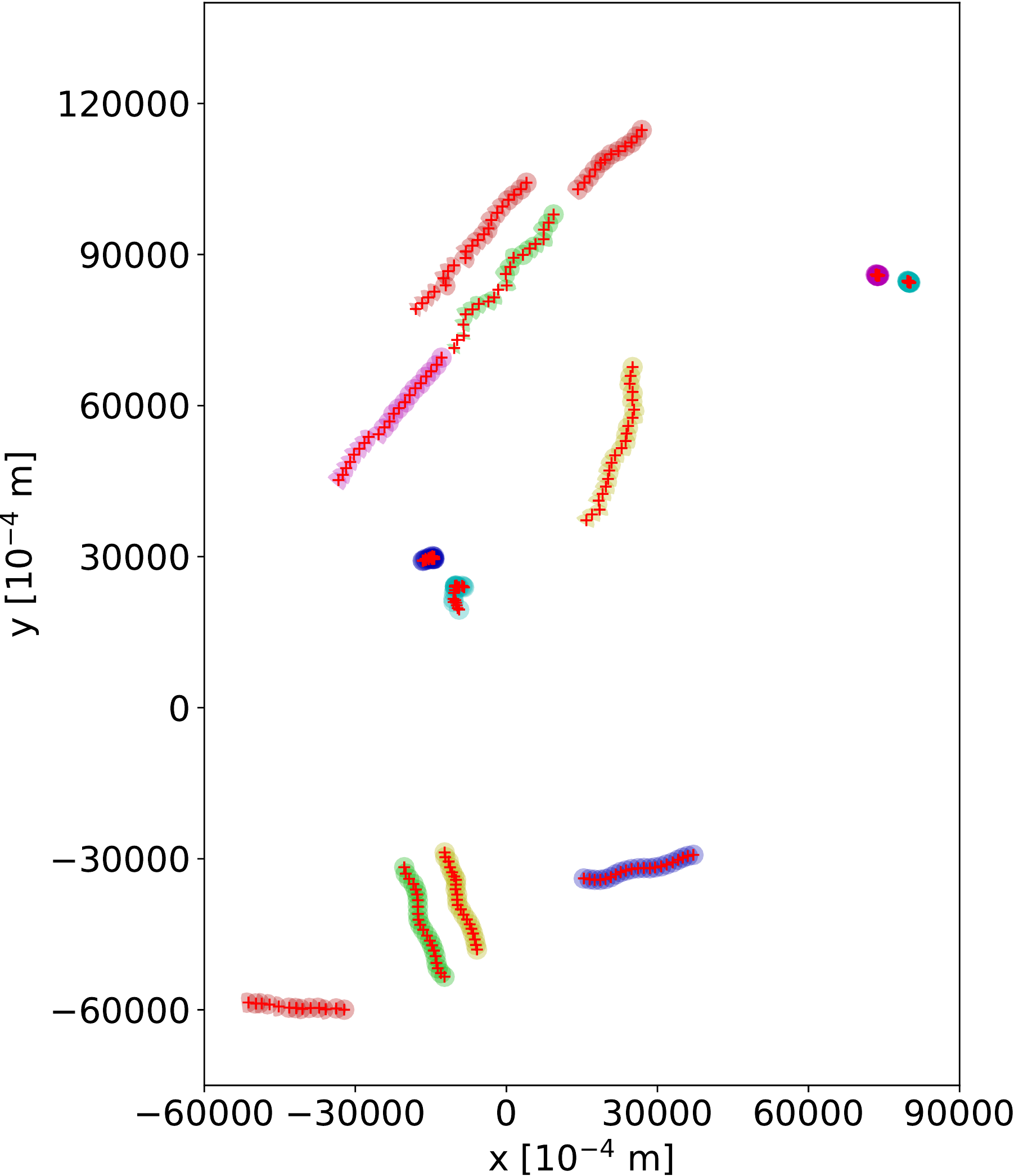} \label{fig:s4}}
\caption{Pedestrian tracking.  \protect\subref{fig:s1} Detection blobs on the image space ($t=0$) estimated by the detector in \cite{pellicano}. Colors refer the estimated height values from $1.4\;m$ (red) to $2\:m$ (green). \protect\subref{fig:s2} Focal elements of the detection BBAs on the ground plane ($t=0$). \protect\subref{fig:s3}Focal elements of track and detection BBAs on the ground plane ($t=8$). Associated tracks and detections share the same color. \protect\subref{fig:s4} Final estimated tracks. Red crosses refer to target locations, while colored sets correspond to regions presenting maximum $BetP$ value. }
\label{fig:frames}
\end{figure*}

\subsubsection{BBA simplification}
A BBA simplification step is essential in tracking applications for two different reasons. First, we want to avoid that the number of focal elements grows without control as the time progresses, because it would mean that the real-time performance of the algorithm would degrade in time, bounding the maximum number of processed frames. Second, we want to avoid an excessive fragmentation of the belief. The BBA simplification aims at reducing the number of focal elements of a given BBA while respecting the least commitment principle. We adopt the method proposed in \cite{ANDRE2015166}, which chooses iteratively two focal elements to aggregate (by performing a union operation) as the ones which minimize the Jousselme's distance \cite{jousselme2001new} between the original BBA and the one obtained after the aggregation.

The proposed representation allows, conversely from the one in \cite{ANDRE2015166} (which simplifies the BBA after each conjunctive combination), to perform the simplification on a less frequent time step. In the proposed experiment a target BBA is simplified when it reaches 15 focal elements, by producing a 5 focal elements BBA. 

\subsubsection{$BetP$ maximization}
At each time step, we run the $BetP$ maximization algorithm presented in Section \ref{betp} for each active track $\widetilde{m}_{t_{i},\delta}$ in order to extract the most probable location of the target. The cardinality of the resulting polygonal set represents the irreducible ambiguity in the target location. The target position is then estimated as the barycenter of the set. 

\subsubsection{Modeling the imprecision of the tracks prediction}
Given the track $\widetilde{m}_{t_{i},\delta}$, which represents the result of the conjunctive combination, we need to model the prediction step imprecision. In order to model the track displacement from the current location, a random walk term is added to the track. Such term boils down to an isotropic dilation of the focal elements. In the proposed representation, this corresponds to applying a scalable polygon offsetting algorithm, having $O(n\log n)$ complexity, where $n$ is the number of vertexes. Polygon offsetting allows a dilation which respects the inclusion relationship of the original focal elements.  The result of such step is the predicted track $m_{t{i},\delta+1}$ at time $t+1$.

\begin{figure} [t]
\captionsetup[subfigure]{labelformat=empty}
\captionsetup[subfloat]{farskip=0pt,captionskip=0.1pt}
\centering
\includegraphics[height=5cm]{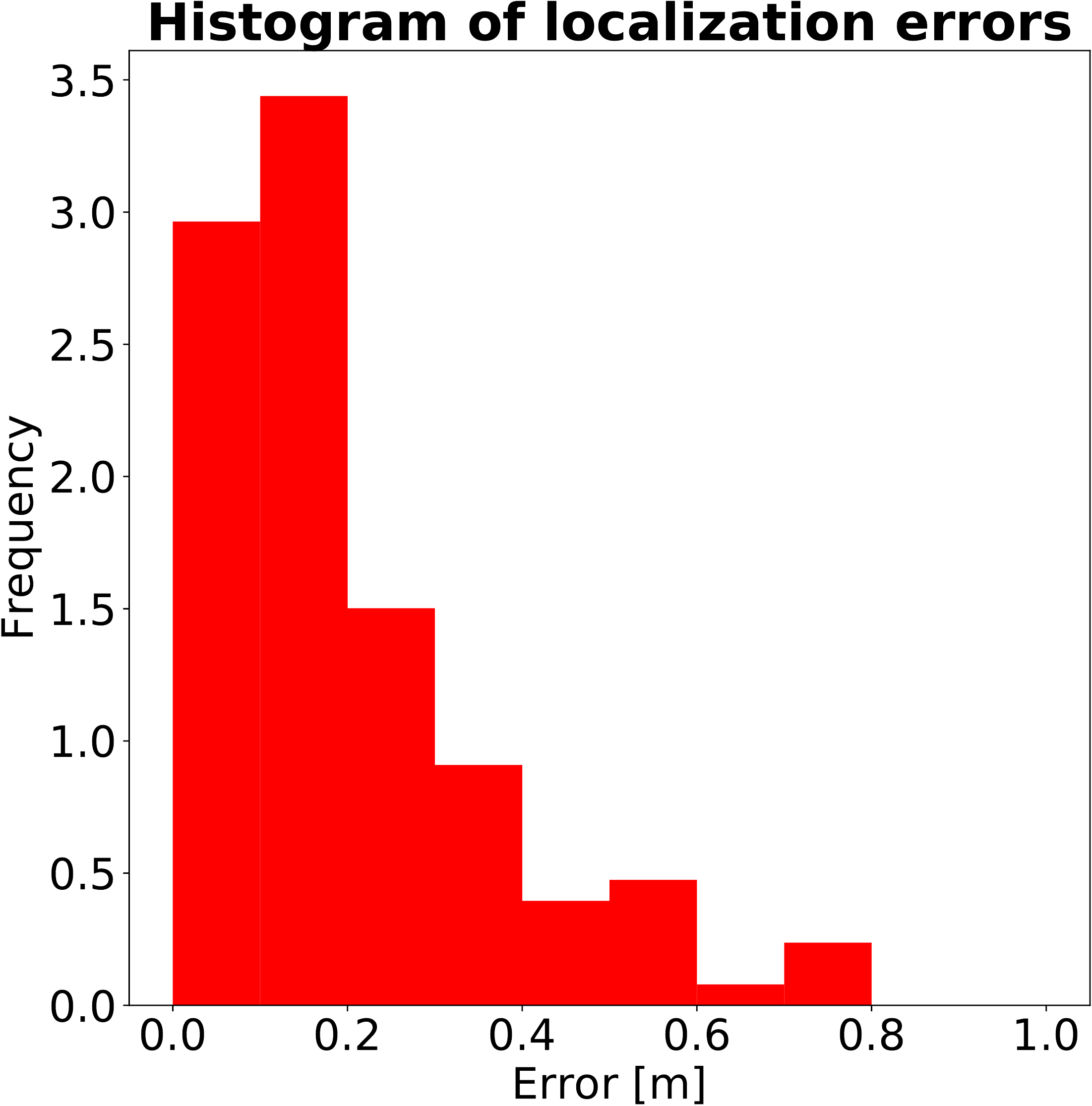}
\caption{Normalized histogram of the localization error of pedestrian tracking on the \textit{Sparse} sequence.}
\label{fig:err}
\end{figure}

\begin{table}
\centering

\begin{minipage}{.49\textwidth}
\centering
\vspace{0.4cm}
\begin{tabular}{|c|c|c|}
\hline
Resolution &  Average Localization error \\
\hline\hline
$10^{-1}\;m$ & $31.061\;cm$  \\
$10^{-2}\;m$ & $22.630\;cm$ \\
$10^{-3}\;m$  & $20.111\;cm$  \\
$10^{-4}\;m$ & $20.006\;cm$  \\
$10^{-5}\;m$ & $20.002\;cm$  \\
\hline
\end{tabular}
\vspace{0.4cm}
\caption{Average localization error on the \textit{Sparse} sequence using different discretization resolutions. By using a representation able to deal with finer resolutions, one may achieve a significant performance gain.}
\label{tab:lbd}
\end{minipage}
\end{table}

\subsubsection{Results}
In order to evaluate quantitatively the tracking accuracy, the target predicted locations are compared against an available ground truth. Such ground truth consists into coordinates in the image space where the heads are located. Since the height of such individuals is not known a priori, each location in the image space projects to a segment in the ground plane, allowing for any possible height in the interval of study. One computes the localization error as the distance between the target estimated location, and the ground truth head location, under the assumption that the height of such head corresponds to the predicted one. Such metric corresponds to computing the distance between the ground truth segment and a height uncertainty segment drawn at the target location. Target locations for inactive track states are estimated by linear regression fit of the estimated target positions at previous states.

Figure \ref{fig:err} shows the results in terms of (normalized) histogram of localization error. The average localization error is $\epsilon=0.2\;m$, which reaches the empiric limit  set by the intrinsic uncertainty of head spatial occupation. On the other hand, the average localization error remains steady in time, meaning that the estimated tracks do not tend to drift away from the real ones. The standard deviation of the average localization error in time is $\sigma=2.3\;cm$.

Table \ref{tab:lbd} shows the average localization error obtained by the tracking algorithm for different choices of the resolution at which the discernment frame is discretized. When a coarse resolution of $10\;cm$ is considered, the performance drops consistently. At this resolution the size of the discernment frame is already large enough to be intractable using methods based on binary representations, as in \cite{ANDRE2015166}. Moreover, while for the theoretically desired resolution of $1\;cm$ the average localization error consistently drops, the proposed representation allows us to scale at finer resolutions to account for rounding errors, thus providing an additional performance boost.  
\section{Conclusion}
This paper proposed a new representation for multi-modal information fusion in bi-dimensional spaces in the BFT domain. Such representation exhibits uniqueness, compactness,  space and precision scalability, which make it suitable for intensive tasks constrained to large hypothesis spaces. We make available a public library for the community, in order to ease the reproducibility of such representation for active research. In our experiments, we show the effectiveness of this formulation on multi-target tracking scenarios, where tenths of tracks have to be estimated on a wide region of interest.

In our future work, we are interested to demonstrate the flexibility of the proposed representation by introducing richer BBAs for detections, in order to model the uncertainty of a detection blob centroid location, which require a non-regular polygon shaping tin order to be exploited. Moreover, we will extend the \textit{2CoBel} library, by studying efficient canonical decomposition approaches.

In terms of application perspectives, we are interested in developing a tracking algorithm for dense crowds, by performing cautious fusion of multiple detection sources from a smart camera network. We aim to demonstrate the use of the proposed representation to make such algorithm scale for high density crowds, for which the number of targets to track jointly can be intractable for state-of-the-art tracking frameworks.


\section*{Acknowledgment}

This work was supported by ANR grant ANR-15-CE39-0005. We gratefully acknowledge the support from Regent's Park Mosque for providing access to the site during the collection of the data used for illustrating our contribution.




\bibliographystyle{IEEEtran}
\bibliography{mybibfile}

%



\end{document}